\documentclass[conference,a4paper]{IEEEtran}

\IEEEoverridecommandlockouts
\usepackage{cite}
\usepackage{amsmath,amssymb}
\usepackage{graphicx}
\usepackage{booktabs}
\usepackage{array}
\usepackage[hidelinks]{hyperref}
\begin{document}
%
\title{Characterizing Energy Consumption in Large Language Models through Design Space Exploration on CPU--GPU Platforms for AI Infrastructure Co-Design}

\author{
\IEEEauthorblockN{Angel Phillips Ortega}
\IEEEauthorblockA{School of Computer Engineering\\
  Technology of Costa Rica\\San Jose, Costa Rica\\
  aphillips@estudiantec.cr}
\and
\IEEEauthorblockN{Luis G. Leon-Vega}
\IEEEauthorblockA{School of Electronics Engineering\\
  Technology of Costa Rica\\San Carlos, Costa Rica\\
  l.leon@itcr.ac.cr}
\and
\IEEEauthorblockN{Niccolo Tosato}
\IEEEauthorblockA{Applied Data Science \& Artificial Intelligence\\
  Universit\`a degli Studi di Trieste\\Trieste, Italy\\
  niccolo.tosato@phd.units.it}
}

\maketitle

\begin{abstract}
Large Language Models (LLMs) increase the computational and energy requirements of artificial-intelligence infrastructure. This work presents a Design Space Exploration (DSE) methodology to characterize energy consumption, generation throughput, and microarchitectural behavior of LLM inference on CPU and GPU platforms. Four models and five GGUF quantizations are evaluated using \texttt{llama.cpp} on a CPU-only node and NVIDIA V100, A100, and H100 nodes. The dataset combines whole-node power monitoring, CPU cache counters, and GPU kernel profiling. Reduced precision improves generation throughput to different extents across GPU generations, while the measured GPU kernel mix remains dominated by matrix operations. Importantly, short accelerator runs and differences in node power envelopes constrain cross-platform energy comparisons. These findings provide a measured baseline for energy-aware AI infrastructure co-design, while distinguishing observed system responses from architectural explanations requiring further validation.
\end{abstract}

\section{Introduction}
Large Language Models (LLMs) have become central to contemporary AI systems, increasing demand for computational throughput and energy-efficient infrastructure. Their deployment relies on platforms combining CPUs and accelerators with different memory organizations and execution characteristics~\cite{ref1_HPC,ref22_HPC}. At the same time, sustainable computing places energy and resource utilization among the primary system design objectives~\cite{ref23_SC}.

Approximate computing and reduced-precision arithmetic can improve computational efficiency~\cite{ref6_ACP}. However, the effect of model choice, quantization, and hardware configuration on measured inference energy is not reducible to a single optimization parameter. In particular, system-level responses alone do not reveal whether observed changes stem from processor execution, memory access, runtime behavior, or the measurement boundary.

This work therefore uses Design Space Exploration (DSE) to characterize LLM inference on CPU and GPU nodes. The objective is \emph{characterization rather than selection of one universally optimal configuration}. The contributions are: (i) a structured DSE workflow linking model, quantization, and platform to measured responses; (ii) a combined system-level and architectural analysis using PSU energy, generation throughput, CPU counters, and GPU kernels; and (iii) an evidence-based discussion of measurement scope and implications for infrastructure co-design. The present evaluation compares CPU-only and GPU-only modes, not simultaneous CPU--GPU partitioning.

\section{Background and Related Work}

DSE evaluates trade-offs among multiple hardware and software design choices~\cite{ref2_DSE}. Prior work surveys heterogeneous AI accelerators and LLM inference mechanisms~\cite{ref1_HPC,ref22_HPC}; other studies consider system energy monitoring in multi-socket GPU
systems~\cite{ref43_efimon,ref44_efimon}. Simulation-based DSE for LLM training systems addresses a related but different problem~\cite{ref4_DSE}. 
Recent studies have also characterized the performance, energy, and quality trade-offs of LLM quantization across GPU architectures~\cite{ref52_llm_quant_characterization}. However, integrating whole node energy measurements, inference throughput, and CPU/GPU architectural profiling within a common DSE framework remains a relevant research opportunity. This work addresses this gap by linking observed system-level responses to CPU cache behavior and GPU kernel activity across different execution platforms. Such a characterization provides complementary architectural insights for energy-aware AI infrastructure co-design, while acknowledging that cross-platform differences do not establish causal effects of individual microarchitectures.

\section{Design Space Exploration Framework}
\subsection{Configurations and Experimental Setup}
The framework maps a design point to the tuple \(d=(h,m,q,e)\), where \(h\) is a node, \(m\) an LLM, \(q\) a GGUF quantization format, and \(e\) an execution mode. Inference is performed using \texttt{llama.cpp}. The evaluated design space comprises four models, five quantization formats, and four nodes (Table~\ref{tab:design}). Quantization formats are \texttt{Q2\_K},
\texttt{Q4\_K\_M}, \texttt{Q5\_K\_M}, \texttt{Q6\_K}, and \texttt{Q8\_0}~\cite{ref42_gguf}. The evaluated models include Phi-3.5-mini-3.8B, Mistral-7B, Llama-3.1-8B, and Gemma-2-9B ~\cite{ref37_Mistral,ref38_llama3,ref39_gemma2}.

The energy dataset comprises 537 executions. The evaluated execution modes are CPU-only and GPU-only. Only observed design-point combinations are analysed; missing combinations are not imputed.


\begin{table}[t]
\centering
\caption{Design space represented in the released dataset.}
\label{tab:design}
\footnotesize
\setlength{\tabcolsep}{3pt}
\begin{tabular}{@{}p{1.04in}p{2.24in}@{}}
\toprule
Dimension & Observed values \\
\midrule
Model & Phi-3.5-mini-3.8B, Mistral-7B, Llama-3.1-8B, Gemma-2-9B \\
Quantization & Q2\_K, Q4\_K\_M, Q5\_K\_M, Q6\_K, Q8\_0 \\
CPU-only node & AMD EPYC baseline (152 energy runs) \\
GPU-only nodes & V100 (143), A100 (128), H100 (114); one accelerator selected per run \\
Runtime & \texttt{llama.cpp}; CPU-only or GPU-only \\
\bottomrule
\end{tabular}
\end{table}

\subsection{Data Acquisition and Processing}

\begin{figure}[ht]
\centering
\includegraphics[width=\columnwidth]{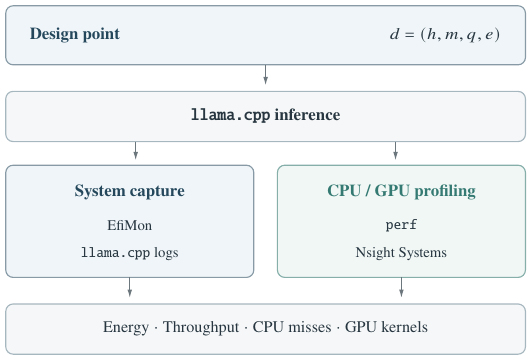}
\caption{Experimental workflow for LLM inference
characterization. Each design point defines an
inference configuration evaluated through system-level
monitoring and architectural profiling to obtain
energy, throughput, and hardware performance metrics.}
\label{fig:data_acquisition}
\end{figure}

Fig.~\ref{fig:data_acquisition} summarizes the experimental workflow, linking each design point to the inference execution, data acquisition, and subsequent characterization.
Each configuration is evaluated using \texttt{llama.cpp}, while EfiMon collects system-level power measurements and execution logs provide inference performance metrics. Complementary architectural profiling is performed using Linux \texttt{perf} for CPU counters and NVIDIA Nsight Systems for GPU kernel activity and memory transfers. The collected data are processed to characterize energy consumption, throughput, and microarchitectural behavior
across the explored design space.

Each observed configuration was executed using \texttt{llama.cpp}. EfiMon monitored whole-node PSU and CPU-socket power at approximately 3-s intervals~\cite{ref43_efimon,ref44_efimon}. Whole-node energy was estimated as $E_{\rm node}=\sum_i P_{{\rm PSU},i}\Delta t_i$; this includes host components and is not isolated GPU energy. Measurements were not corrected for idle power. Table~\ref{tab:coverage} summarizes monitoring coverage; short GPU runs often contain only one power sample, making their estimated joules sensitive to sampling alignment. $N_s$ denotes the number of samples, and $T_{\rm EfiMon}$ is the sampled window, not the \texttt{llama.cpp} inference time.

\begin{table}[t]
\centering
\caption{Energy-data coverage and temporal resolution.}
\label{tab:coverage}
\footnotesize
\setlength{\tabcolsep}{3pt}
\begin{tabular}{@{}lrrrr@{}}
\toprule
Node & Runs & Median $N_s$ & Median $T_{\rm EfiMon}$ (s) & $N_s=1$ \\
\midrule
CPU/EPYC & 152 & 91 & 307.92 & 0 \\
V100 & 143 & 2 & 5.76 & 69 \\
A100 & 128 & 1 & 2.88 & 112 \\
H100 & 114 & 1 & 3.04 & 102 \\
\bottomrule
\end{tabular}
\end{table}

Generation throughput and token counts were obtained from \texttt{llama.cpp} logs. Linux \texttt{perf} supplied CPU cache counters, while NVIDIA Nsight Systems supplied GPU kernel, CUDA API, and transfer activity~\cite{ref45_nsight_systems}. The profiling exports include 456 GPU runs, 152 AMD CPU runs, and a separate Intel dataset of 95 observations; profiling and energy records are not time-aligned. Prompts contain 106--125 tokens, generated lengths range from 156 to 419 tokens, and configuration repetitions are typically 5--8. For each observed node/model/quantization combination, we report run medians and within-node Q4/Q8 ratios; no missing configurations are imputed. Exact prompt texts, runtime revision, command-line options, and sensor calibration are not documented in the released tables, limiting independent reproduction.

\subsection{Response and Architectural Metrics}
System-level responses are whole-node energy (J), execution time (s), and generation throughput (tokens/s). Token/J is not used to rank GPUs because the measurement boundary is the full node. Architectural responses are CPU L1/L2/LLC miss ratios, GPU kernel-time composition, and CUDA transfer and runtime activity. Comparisons between platforms are observational because host topology, memory placement, and thread settings are not controlled across nodes.

\section{Results and Analysis}
\subsection{System-Level Energy and Performance}
\begin{figure}[t]
\centering
\includegraphics[width=\columnwidth]{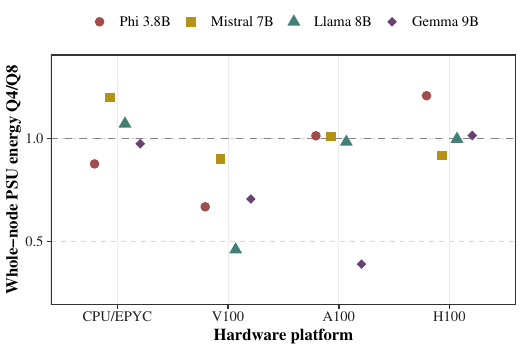}
\caption{Q4\_K\_M/Q8\_0 ratio of median whole-node PSU energy, within each node and model. One point denotes one model. Ratios are \emph{not} GPU-only energy-efficiency comparisons; short runs have sparse power samples.}
\label{fig:energy}
\end{figure}

Fig.~\ref{fig:energy} compares Q4 to Q8 using the ratio of the median whole-node PSU energy for each node/model pair. For V100, Q4/Q8 ratios across the four models range from 0.46 to 0.90; for A100, most are near 1.0, with Gemma-2-9B an exception (0.39); and for H100, they range from 0.92 to 1.21. The CPU-only node also shows mixed changes (0.88--1.20). These are descriptive short-window measurements; the unusually low A100/Gemma ratio (0.39) requires calibrated replication before being treated as an energy gain.

\begin{figure}[t]
\centering
\includegraphics[width=\columnwidth]{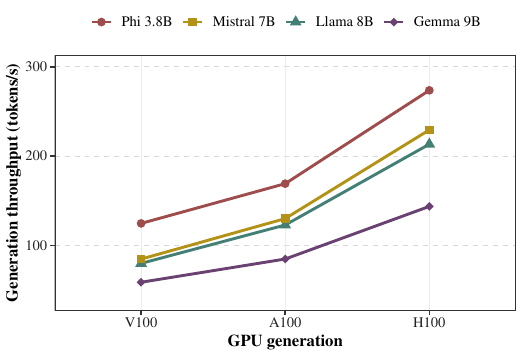}
\caption{Median \texttt{llama.cpp} generation throughput under Q8\_0; different architectures also differ in their host configurations and measured output lengths.}
\label{fig:tps}
\end{figure}

Model selection and quantization also shape generation throughput. For Q8, the V100, A100, and H100 achieve respectively 79.93, 122.91, and 213.38 tokens/s on Llama-3.1-8B (Fig.~\ref{fig:tps}). Across matched model/quantization settings, the median A100/V100 and H100/V100 throughput ratios are 1.37 and 2.20. Reducing Q8 to Q4 improves throughput by a median factor of 1.37 on V100, 1.26 on A100, 1.16 on H100, and 1.20 on the CPU baseline (each median taken across the four models). Thus, quantization remains beneficial for \emph{throughput} in these runs, but its magnitude is platform-dependent; a corresponding inference-quality change was not evaluated.

\subsection{Impact of Hardware Configuration}
\textit{CPU microarchitectures.} The AMD baseline exhibits L1-data miss ratios generally around 2--6\%, while the separately supplied Intel profiling export reports approximately 14--20\% for similar quantization levels. L2 miss ratios and counter semantics also differ. These observations are consistent with distinct memory-hierarchy behavior, but the Intel and AMD systems do not have controlled core count, thread, memory, and topology settings. We therefore do not attribute the original order-of-magnitude PSU-energy differences to vendor or cache misses; the available release does not establish a matched CPU-energy comparison.

\begin{figure}[t]
\centering
\includegraphics[width=\columnwidth]{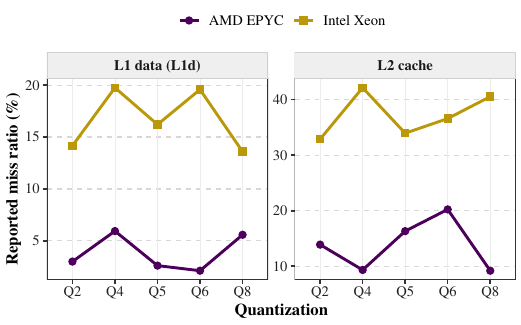}
\caption{Median cache-miss ratios from distinct AMD and Intel profiling datasets. The CPU platforms and counter definitions are not fully matched; differences are observational, not a vendor causal effect.}
\label{fig:cache}
\end{figure}

Fig.~\ref{fig:cache} separates L1-data and L2 observations instead of compressing all cache levels into a small multi-panel grid. The markedly different reported last-level-cache values in the exports also warrant checking event definitions and counter multiplexing before a direct vendor comparison. This is especially important because a miss percentage is not itself an energy measurement: cache behavior may affect stall time, but the magnitude of any energy effect depends on the memory subsystem and the execution-time/power trade-off. Thus, the original observation about memory-hierarchy sensitivity remains a hypothesis for targeted experimentation, rather than an explanation for a large measured energy gap.

\begin{figure}[t]
\centering
\includegraphics[width=\columnwidth]{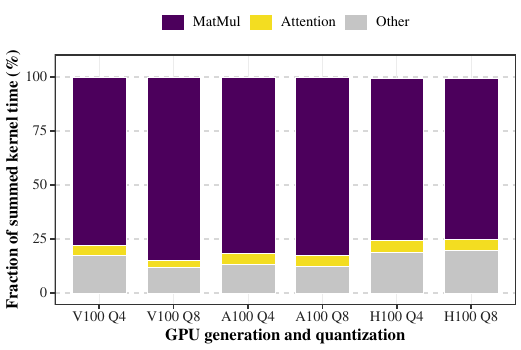}
\caption{Median share of summed CUDA-kernel time, grouped by GPU and quantization. ``MatMul'' includes \texttt{mul\_mat} and \texttt{cutlass::Kernel} name matches; classification is descriptive.}
\label{fig:kernels}
\end{figure}

\textit{GPU microarchitectures.} The GPU profiles reveal a consistent computational hotspot (Fig.~\ref{fig:kernels}). Across GPU generations and precisions, kernels identified as matrix multiplication account for the largest fraction of summed kernel time: approximately 74--85\% in the Q4/Q8 grouping shown. The mix varies with quantization and generation, but kernel totals do not include all host, loading, and synchronization overheads. For example, Llama-3.1-8B/Q8 summed kernel times are 3.12 s (V100), 0.062 s (A100), and 0.034 s (H100); these differences \emph{must not} be reported as end-to-end 50--100$\times$ speedups. The observed generation-throughput comparison is much smaller, as given above. Version-dependent kernel paths may also contribute and require further isolation.

\subsection{Data Intensity and Transfer Activity}
The profiling index relating host--device transfer volume to GPU kernel time is
\begin{equation}
I_{\rm transfer}=\frac{V_{\rm HtoD}+V_{\rm DtoH}}{T_{\rm CUDA,kernels}},
\label{eq:intensity}
\end{equation}
where volumes and kernel durations are reported by Nsight. Although expressed in bytes/s, this quotient is not sustained memory bandwidth: transfers and kernels are distinct activities that may overlap in wall-clock time.

\begin{table}[t]
\centering
\caption{GPU profiling index for Llama-3.1-8B (median of runs).}
\label{tab:intensity}
\footnotesize
\setlength{\tabcolsep}{3pt}
\begin{tabular}{@{}lrrrr@{}}
\toprule
& \multicolumn{2}{c}{$I_{\rm transfer}$ (GB/s)} & \multicolumn{2}{c}{Kernel time (s)} \\
GPU & Q4 & Q8 & Q4 & Q8 \\
\midrule
V100 & 1.95 & 2.60 & 2.442 & 3.124 \\
A100 & 79.72 & 130.69 & 0.060 & 0.062 \\
H100 & 137.30 & 240.39 & 0.035 & 0.034 \\
\bottomrule
\end{tabular}
\end{table}

Table~\ref{tab:intensity} shows that Q4 can reduce the transfer-per-kernel-time index even when generation throughput increases. Normalizing by short kernel durations also amplifies cross-generation differences. Because power and profiling measurements come from separate executions, this index neither establishes equivalent inference performance nor supports comparisons of GPU energy efficiency.

\subsection{Design Space Interpretation and Limitations}
The DSE combines three complementary perspectives: energy trends within a node, end-to-end generation throughput, and internal processor activity. The results show that the value of lowering numerical precision depends on hardware and on the response being optimized: throughput may improve without a reliable, proportional reduction in measured whole-node energy. Kernel composition identifies a plausible target for further optimization, while CPU cache statistics suggest memory locality as a variable worth isolating experimentally.

The principal threats to validity are the PSU measurement boundary, sparse samples on short GPU runs, unequal host architectures and memory configurations, non-fixed output lengths, and incomplete design-point coverage. Profiling and energy data are separate runs rather than temporally aligned phase measurements. The work does not quantify inference quality, mixed CPU--GPU layer offload, or GPU-specific joules. Accordingly, these results support comparative \emph{characterization}, not a causal ranking of accelerator generations by energy efficiency.

\begin{table}[t]
\centering
\caption{Evidence-based interpretation of the explored DSE.}
\label{tab:summary}
\footnotesize
\setlength{\tabcolsep}{3pt}
\begin{tabular}{@{}l>{\raggedright\arraybackslash}p{1.09in}>{\raggedright\arraybackslash}p{1.41in}@{}}
\toprule
Dimension & Measured response & Co-design implication \\
\midrule
Model & Throughput differs by model & Match architecture to workload \\
Precision & Q4 throughput benefits vary & Assess per device; verify quality \\
CPU cache & Unmatched miss-rate profiles & Test locality with controls \\
GPU kernels & Matrix operations dominate & Profile backend paths \\
Energy & Whole-node PSU; sparse GPU samples & Calibrate before ranking GPUs \\
\bottomrule
\end{tabular}
\end{table}

Table~\ref{tab:summary} replaces the earlier ordinal factor ranking with evidence-based implications. Precision can change memory footprint, dequantization, and kernel selection together, while GPU-generation differences are confounded with host configuration. The results therefore support platform-level associations rather than isolated microarchitectural causes. A controlled follow-up should balance design points, fix workload lengths and thread affinity, and capture sufficiently sampled energy traces.

\section{Conclusions and Future Work}
This paper presented a DSE methodology to characterize LLM energy, throughput, and microarchitectural response on CPU and GPU platforms. The measurements indicate architecture-dependent throughput benefits from quantization and continued dominance of matrix-operation kernels in GPU profiling. Distinguishing whole-node energy from per-accelerator power, and profiler kernel time from end-to-end inference time, is essential for reliable co-design conclusions. Future work will calibrate energy capture for short runs, control workload and host confounders, incorporate output-quality evaluation, and extend the design space to mixed CPU--GPU execution and additional approximation methods. With these changes, our work will help future computing system dimensioning align with applications and workloads, enabling more informed decisions based on application requirements.

\section*{Acknowledgment}
The authors used AI-assisted tools to support manuscript editing and the development of figure-generation scripts. The authors reviewed the analyses and remain responsible for the accuracy of the results, interpretations, and conclusions.

This work was supported by the Costa Rica Institute of Technology under the research project 1360058 (Generación Automática de Hardware para Aplicaciones de Aprendizaje Automático basadas en FPGA). Also, the results were thanks to the access to the ORFEO supercomputer provided by AREA Science Park.


\clearpage

\bibliographystyle{IEEEtran}
\bibliography{references}

@article{ref1_HPC,
author = {Cristina Silvano and others},
title = {A Survey on Deep Learning Hardware Accelerators for Heterogeneous HPC Platforms},
year = {2025},
issue_date = {November 2025},
publisher = {Association for Computing Machinery},
address = {New York, NY, USA},
volume = {57},
number = {11},
issn = {0360-0300},
url = {https://doi.org/10.1145/3729215},
doi = {10.1145/3729215},
journal = {ACM Comput. Surv.},
month = jun,
articleno = {286},
numpages = {39}
}

@ARTICLE{ref2_DSE,
  author={Pimentel, Andy D.},
  journal={IEEE Design \& Test}, 
  title={Exploring Exploration: A Tutorial Introduction to Embedded Systems Design Space Exploration}, 
  year={2017},
  volume={34},
  number={1},
  pages={77-90},
  doi={10.1109/MDAT.2016.2626445}
}

@inproceedings{ref4_DSE,
  title={Accelerating Design Space Exploration for LLM Training Systems with Multi-experiment Parallel Simulation},
  author={Fei Gui and Kaihui Gao and Li Chen and Dan Li and Vincent Liu and Ran Zhang and Hongbing Yang and Dian Xiong},
  booktitle={Symposium on Networked Systems Design and Implementation},
  year={2025},
  url={https://api.semanticscholar.org/CorpusID:278205155}
}

@ARTICLE{ref6_ACP,
  author={Xu, Qiang and Mytkowicz, Todd and Kim, Nam Sung},
  journal={IEEE Design \& Test}, 
  title={Approximate Computing: A Survey}, 
  year={2016},
  volume={33},
  number={1},
  pages={8-22},
  doi={10.1109/MDAT.2015.2505723}
}

@misc{ref22_HPC,
      title={Large Language Model Inference Acceleration: A Comprehensive Hardware Perspective}, 
      author={Jinhao Li and Jiaming Xu and Shan Huang and Yonghua Chen and Wen Li and Jun Liu and Yaoxiu Lian and Jiayi Pan and Li Ding and Hao Zhou and Yu Wang and Guohao Dai},
      year={2025},
      eprint={2410.04466},
      archivePrefix={arXiv},
      primaryClass={cs.AR},
      url={https://arxiv.org/abs/2410.04466}, 
}

@article{ref23_SC,
author = {Bashir, Noman and Irwin, David and Shenoy, Prashant and Souza, Abel},
title = {Sustainable Computing - Without the Hot Air},
year = {2023},
issue_date = {October 2023},
publisher = {Association for Computing Machinery},
address = {New York, NY, USA},
volume = {3},
number = {3},
url = {https://doi.org/10.1145/3630614.3630623},
doi = {10.1145/3630614.3630623},
journal = {SIGENERGY Energy Inform. Rev.},
month = oct,
pages = {47–52},
numpages = {6}
}

@article{ref37_Mistral,
  title        = {Mistral 7B},
  author       = {Jiang, Albert Q. and others},
  journal      = {arXiv preprint arXiv:2310.06825},
  year         = {2023},
  doi          = {10.48550/arXiv.2310.06825},
  url          = {https://arxiv.org/abs/2310.06825}
}

@article{ref38_llama3,
  title        = {The Llama 3 Herd of Models},
  author       = {Dubey, Abhimanyu and Jauhri, Abhinav and Pandey, Abhinav and Kadian, Abhishek and Al-Dahle, Ahmad and Letman, Aiesha and Mathur, Akhil and Schelten, Alan and Yang, Amy and Fan, Angela and others},
  journal      = {arXiv preprint arXiv:2407.21783},
  year         = {2024},
  doi          = {10.48550/arXiv.2407.21783},
  url          = {https://arxiv.org/abs/2407.21783}
}

@article{ref39_gemma2,
  title        = {Gemma 2: Improving Open Language Models at a Practical Size},
  author       = {{Gemma Team}},
  journal      = {arXiv preprint arXiv:2408.00118},
  year         = {2024},
  doi          = {10.48550/arXiv.2408.00118},
  url          = {https://arxiv.org/abs/2408.00118}
}

@misc{ref42_gguf,
  title        = {GGUF File Format Specification},
  author       = {GGML and llama.cpp Contributors},
  year         = {2024},
  howpublished = {\url{https://github.com/ggml-org/ggml/blob/master/docs/gguf.md}}
}

@InProceedings{ref43_efimon,
author="Le{\'o}n-Vega, Luis G.
and Tosato, Niccol{\`o}
and Cozzini, Stefano",
editor="Guerrero, Gin{\'e}s
and San Mart{\'i}n, Jaime
and Meneses, Esteban
and Barrios Hern{\'a}ndez, Carlos Jaime
and Osthoff, Carla
and Monsalve Diaz, Jose M.",
title={{A Comprehensive Analysis of Process Energy Consumption on Multi-socket Systems with GPUs}},
booktitle="High Performance Computing",
year="2025",
publisher="Springer Nature Switzerland",
address="Cham",
pages="52--67",
isbn="978-3-031-80084-9"
}

@InProceedings{ref44_efimon,
author="Le{\'o}n-Vega, Luis G.
and Tosato, Niccol{\`o}
and Cozzini, Stefano",
editor="Guerrero, Gin{\'e}s
and San Mart{\'i}n, Jaime
and Meneses, Esteban
and Barrios Hern{\'a}ndez, Carlos Jaime
and Osthoff, Carla
and Monsalve Diaz, Jose M.",
title="EfiMon: A Process Analyser for Granular Power Consumption Prediction",
booktitle="High Performance Computing",
year="2025",
publisher="Springer Nature Switzerland",
address="Cham",
pages="112--126",
isbn="978-3-031-80084-9"
}

@manual{ref45_nsight_systems,
  title        = {NVIDIA Nsight Systems User Guide},
  organization = {NVIDIA Corporation},
  year         = {2025},
  url          = {https://docs.nvidia.com/nsight-systems/UserGuide/},
  note         = {Accessed: 2026-08-02}
}

@misc{ref52_llm_quant_characterization,
  title = {Systematic Characterization of LLM
           Quantization: A Performance, Energy,
           and Quality Perspective},
  author = {Shi, Tianyao and Ding, Yi},
  year = {2025},
  eprint = {2508.16712},
  archivePrefix = {arXiv},
  primaryClass = {cs.DC},
  url = {https://arxiv.org/abs/2508.16712}
}

\end{document}